\documentclass[final]{cvpr}
\newcommand{\final}{1}

\usepackage{times}
\usepackage{epsfig}
\usepackage{graphicx}
\usepackage{amsmath}
\usepackage{amssymb}
\usepackage{amsthm}
\usepackage{subcaption}
\usepackage{url}
\usepackage{gensymb}
\usepackage{ulem}
\usepackage[symbol]{footmisc}
\ifdefined\siggraph
\usepackage{times}
\fi

\usepackage{color}
\usepackage{ifthen}
\usepackage{float}
\usepackage{alltt}
\usepackage{newlfont} 
\usepackage{wrapfig}
\usepackage{booktabs}
\usepackage{multirow}
\usepackage{array}
\usepackage{appendix}
\usepackage{comment}
\usepackage{amsmath}
\usepackage{amssymb}
\usepackage{amsthm}
\usepackage{textcomp}

\newcommand{\nothing}[1]{}

\definecolor{DeltaColor}{rgb}{0.039,0.73,0.71}
\definecolor{SetaColor}{rgb}{0.867, 0.0235, 0.376}
\definecolor{SigmaColor}{rgb}{0.98,0.45,0.0}
\definecolor{RedColor}{rgb}{0.8,0,0}
\definecolor{AlphaColor}{rgb}{0,0,0.8}
\definecolor{BetaColor}{rgb}{0.8,0,0.8}
\definecolor{GammaColor}{rgb}{0.5,0,0.7}
\definecolor{EpsilonColor}{rgb}{0.353,0.725,0.906}
\definecolor{TauColor}{rgb}{0.423,0.235,0.192}
\newcommand{\weikai}[1]{{\color{RedColor} Weikai: #1 $\qed$}}
\newcommand{\yuda}[1]{{\color{AlphaColor} Yuda: #1 $\qed$}}
\newcommand{\xiaojie}[1]{{\color{SigmaColor} Xiaojie: #1 $\qed$}}
\newcommand{\xiaoguang}[1]{{\color{GammaColor} Xiaoguang: #1 $\qed$}}
\newcommand{\yushuang}[1]{{\color{DeltaColor} Yushuang: #1 $\qed$}}

\definecolor{AudioColor}{rgb}{0.56,0.34,0.62}

\definecolor{DeadlineColor}{rgb}{0.9,0.4,0} 

\definecolor{figred}{rgb}{1,0,0}
\definecolor{figgreen}{rgb}{0,0.6,0}
\definecolor{figblue}{rgb}{0,0,1}
\definecolor{figpink}{rgb}{1,0.63,0.63}

\newcolumntype{C}[1]{>{\centering}m{#1}}

\ifthenelse{\equal{\final}{1}}
{
\renewcommand{\yuda}[1]{}
\renewcommand{\weikai}[1]{}
\renewcommand{\xiaojie}[1]{}
\renewcommand{\xiaoguang}[1]{}
}
{}

\newcounter{pccount}
\floatstyle{plain}

\newcommand{\filename}[1]{\url{#1}}
\newcommand{\foldername}[1]{\url{#1}}

\newcommand{\dname}{3DCaricShop} 
\newcommand{\meshTemp}{\mathbf{M_t}} 
\newcommand{\meshPIFu}{\mathbf{M_I}} 
\newcommand{\meshNICP}{\mathbf{M_N}} 
\newcommand{\meshPCA}{\mathbf{M_P}} 

\usepackage[misc]{ifsym}
\usepackage[pagebackref=true,breaklinks=true,colorlinks,bookmarks=false]{hyperref}

\def\cvprPaperID{2456} 
\def\confYear{CVPR 2021}

\graphicspath{
	{figs/raster/}
	{figs/handdrawn/}
}

\begin{document}
	%
	\title{3DCaricShop: A Dataset and A Baseline Method for Single-view \\3D Caricature Face Reconstruction}
	
    \author{Yuda Qiu\textsuperscript{1} \qquad 
    Xiaojie Xu\textsuperscript{1} \qquad 
    Lingteng Qiu\textsuperscript{1} \qquad 
    Yan Pan\textsuperscript{1}\\
    Yushuang Wu\textsuperscript{1} \qquad
    Weikai Chen\textsuperscript{2} \qquad
    Xiaoguang Han\textsuperscript{1,}\footnotemark[1]\\
    
    \textsuperscript{1}SRIBD, The Chinese University of Hong Kong, Shenzhen\footnotemark[2] \qquad
    \textsuperscript{2}Tencent America\\
    }

	\maketitle
    \footnotetext[1]{\textbf{Corresponding email:} hanxiaoguang@cuhk.edu.cn}	
    \footnotetext[2]{Shenzhen Research Institute of Big Data}
	\begin{abstract}

Caricature is an artistic representation that deliberately exaggerates the distinctive features of a human face to convey humor or sarcasm. However, reconstructing a 3D caricature from a 2D caricature image remains a challenging task, mostly due to the lack of data. We propose to fill this gap by introducing \dname{}, the first large-scale 3D caricature dataset that contains 2000 high-quality diversified 3D caricatures manually crafted by professional artists. 
\dname{} also provides rich annotations including a paired 2D caricature image, camera parameters and  3D facial landmarks. To demonstrate the advantage of \dname{}, we present a novel baseline approach for single-view 3D caricature reconstruction. To ensure a faithful reconstruction with plausible face deformations, we propose to connect the good ends of the detail-rich implicit functions and the parametric mesh representations. In particular, we first register a template mesh to the output of the implicit generator and iteratively project the registration result onto a pre-trained PCA space to resolve artifacts and self-intersections. To deal with the large deformation during non-rigid registration, we propose a novel view-collaborative graph convolution network (VC-GCN) to extract key points from the implicit mesh for accurate alignment. Our method is able to generate high-fidelity 3D caricature in a pre-defined mesh topology that is animation-ready. Extensive experiments have been conducted on \dname{} to verify the significance of the database and the effectiveness of the proposed method. We will release \dname{} upon publication.
\end{abstract}

\nothing{
Caricature is an exaggerate representation for human face, which highlights the personal features. 
Due to the lack of 3D caricature data, modelling 3d shape from caricature images with large diversity of geometric and texture variation is a challenging task. 
In this paper, we propose the first 3d caricature dataset built by artists, with various styles of images and the corresponding 3d shapes. Based on the dataset, we propose a baseline approach, combining implicit and parametric method together, to reconstruct face mesh with uniform topology. To solve the registration problem between two corpus meshes robustly, we further propose a novel 3d landmark detection algorithm for meshes from implicit method. Experiments show the effectiveness of our dataset and the proposed baseline method.  
}
	\begin{figure}
   \begin{center}
    \includegraphics[width=3in]{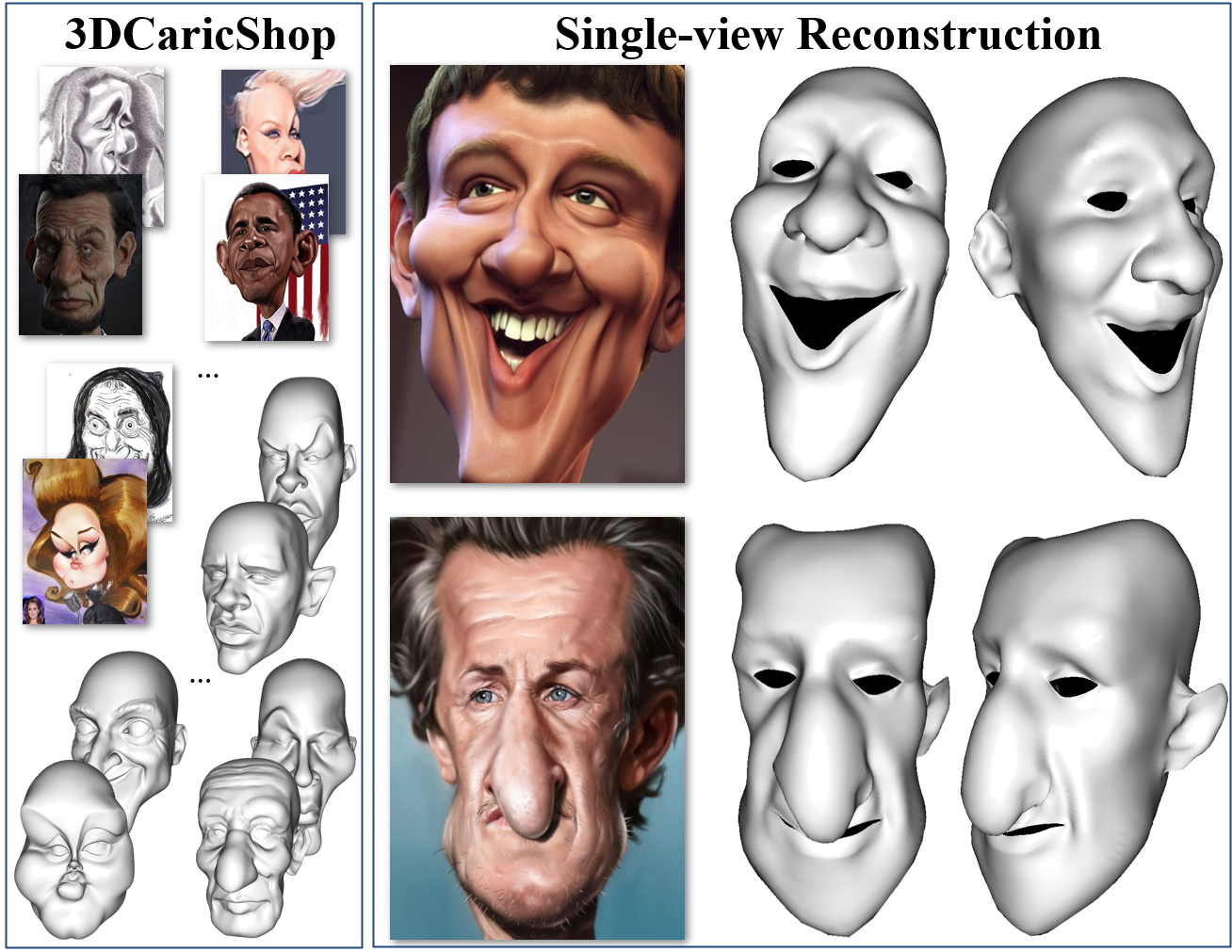}
   \end{center}
   \vspace{-4mm}
  \caption{{\bf Left}: the proposed \dname{}, a large-scale repository of 3D caricatures that are manually crafted by professional artists. \dname{} is richly annotated with 2D caricature images, camera parameters, and 3D facial landmarks. {\bf Right}: the proposed baseline method that sets the new state of the art in single-view 3D caricature reconstruction.}
  \label{fig:teaser}
  \vspace{-4mm}
\end{figure}
	\section{Introduction}
\label{sec:intro}


A caricature is a vivid art form of depicting persons by abstracting or exaggerating the peculiarities of the facial features. 
As a way to convey humor or sarcasm, caricatures are widely used in entertainment, social events, electronic games and a variety of artistic creations.
While 2D caricatures have gained popularity in comic graphics, there exist many scenarios, including cartoon character creation, game avatar customization, custom-made 3D printing, etc., that the 3D face caricatures remain the mainstream representations. 
However, creating a high-quality 3D caricature is a labor-intensive and time-consuming task even for a skilled artist.
Thereby, generating expressive 3D face caricatures from a minimal input, such as a single image, is a highly-demanding but also challenging task.


Most of the prior works mainly focus on 2D caricature generation~\cite{cao2018carigans, shi2019warpgan, Kim20DST}, while research on reconstructing 3D caricatures from 2D caricature images remains vary rare.
Wu et al.~\cite{wu2018alive} propose the first work that creates 3D caricature from 2D caricature images using an optimization based approach.
They formulate caricature generation as a problem of deforming the standard 3D face. 
In particular, they build a intrinsic deformation space based on the exaggerated morphable models of standard faces~\cite{paysan20093d}.
The deformation coefficients are then optimized to reduce the landmark fitting errors.
Recently, in their follow-up work~\cite{zhang2020landmark}, they employ CNN to automate the task of 2D facial landmark prediction and deformation regression.
However, previous works~\cite{sela2017unrestricted,liu20193d} have shown that the traditional 3D morphable models (3DMM) of normal faces have very limited expressiveness in modeling the intricate facial deformations in reality.
Thereby, the deformation space based on a synthetically exaggerated 3DMM, as proposed in \cite{wu2018alive,zhang2020landmark}, is far from sufficient to capture realistic 3D caricatures, which are even more diversified and complex than normal faces.

The key to tackling the above problem is a high-quality 3D caricature dataset created by artists that can provide realistic shape priors for both learning-based and optimization-based approaches. 
However, there exist two challenges in constructing such a dataset.
First, the 3D models crafted by artists are not topologically consistent, making it infeasible to many downstream applications, including blendshape creation, face animation, 3D landmark localization, etc. 
Secondly, the manually created meshes are typically not aligned with the corresponding images.
While many face reconstruction techniques require an accurate registration, such misalignment makes the dataset inapplicable to projection-based applications such as landmark fitting, texture restoration and manipulation, etc.


In this work, we introduce \dname{}, a large-scale 3D caricature dataset that simultaneously addresses the above issues.
First of all, \dname{} contains 2,000 highly diversified and high-quality 3D caricature models manually crafted by professional artists. 
It is constructed by requesting artists to create 3D caricatures according to 2,000 manually selected caricature images from WebCaricature~\cite{huo2017webcaricature}, that span a wide range of shape exaggerations and texturing styles.
Compared to the synthetic datasets~\cite{han2017deepsketch2face,zhang2020landmark}, \dname{} can provide shape priors for 3D caricatures with much higher fidelity.
Secondly, all the 3D models in \dname{} have been re-topologized to a consistent mesh topology that paves the way to a number of future applications, including learning a parametric shape space, batch geometry processing, etc. 
Thirdly, we provide accurate 3D face landmarks in \dname{}, which facilitates the use of landmark fitting technique that is widely adopted in the state-of-the-art face reconstruction approaches.
Last but not least, \dname{} offers a paired 2D caricature image and the camera parameters that are used for mesh alignment.
This enables a wide range of techniques, such as differentiable rendering, landmark fitting, etc., that rely on 2D-to-3D consistency. 

To further exploit the power of \dname{}, we propose a novel baseline approach to infer 3D caricatures from a single caricature image.
While the methods based on deep implicit functions~\cite{huang2018deep,park2019deepsdf} have shown promising capability of modeling objects with arbitrary topologies,
it is prone to artifacts and self-intersections when applied to reconstruct 3D caricatures, which typically contain many extreme distortions. 
Though approaches using parametric mesh model can ensure a generation of plausible 3D face, they struggle to produce realistic faces with accurate geometry.
We advocate to connect the good ends of both worlds by transferring the high-fidelity geometry learnt from the implicit reconstruction to a template mesh with a reasonable topology.
To enable a faithful transfer, we propose a novel view-collaborative graph convolution network (VC-GCN) to extract key points from the implicit mesh for accurate mesh alignment. 
To strike a balance between accuracy and robustness, we iteratively project the registered template mesh onto a pre-trained PCA space using \dname{} to avoid overfitting to outliers.
Our approach is able to generate high-quality 3D caricatures in a pre-defined mesh topology that is animation-ready.

We have conducted extensive benchmarking and ablation analysis on the proposed dataset. 
Experimental results show that the proposed approach trained on \dname{} sets new state of the art on the task of single-view 3D caricature reconstruction from caricature images. 
\nothing{
Our contributions are summarized as follows:
\vspace{-2mm}
\begin{itemize}
	\item We build 3DCaricShop, the first large-scale, richly annotated 3D caricature dataset that contains 2,000 high-quality 3D caricature models crafted by professional artists.
	\vspace{-2mm}
	\item We present a novel baseline method that can faithfully reconstruct a 3D caricature model from a single caricature image with uniform topology. 
	\vspace{-2mm}
	\item We propose a novel approach to detect 3D landmarks from the meshes reconstructed using deep implicit functions.
\end{itemize}
}

\nothing{

\weikai{Compile draft.tex to generate draft with comments. Compile submission.tex to generate draft without comments, which is ready for submission.}

\yuda{Test!}

\yushuang{Test!}

\xiaojie{Test!}

\xiaoguang{Test!}
}

	\section{Related Work}
\label{sec:related_work}

\paragraph{Single-view Reconstruction} 
Single-view reconstruction (SVR) is a classic task in computer vision. Existing methods could be classified as reconstruction for general objects \cite{groueix2018papier,wang2018pixel2mesh,park2019deepsdf} and for objects in specific categories\cite{cao2013facewarehouse,bogo2016keep,zuffi2018lions}. 
It is an ill-posed problem due to the ambiguous nature. In tradition, strong priors are introduced to constrain the space of solutions. Shape-from-Shading (SfS)~\cite{prados2006shape} is a kind of physical based prior on the relation between illumination and shape, which recovers the detailed shape in photos. However, it fails to analyze artists works because of the stylized shading effect.
Most recently, with the success of deep learning architectures and the release of large-scale 3D shape datasets such as ShapeNet~\cite{chang2015shapenet}, learning based approaches have achieved great progress, by learning the shape priors directly from the huge datasets. According to the used 3D representations, these methods can be divided into voxel-based~\cite{maturana2015voxnet,choy20163d,Riegler_2017_CVPR}, point-based~\cite{qi2017pointnet,qi2017pointnet++}, mesh-based~\cite{wang2018pixel2mesh,pan2019deep}, and implicit-function-based frameworks~\cite{mescheder2019occupancy, saito2019pifu}. Among these methods, PIFu~\cite{saito2019pifu}, an algorithm based on implicit functions, has been applied on the reconstruction of human body and achieves impressive results. In this paper, we employ PIFu to create the 3D mesh for each single caricature image.

\vspace{-4mm}
\paragraph{Single-view Face Modeling}
A closely-related task is photo-realistic face reconstruction. Two mainstream methodologies are developed to handle this problem, i.e. parametric based~\cite{cao2013facewarehouse,tran2018nonlinear,deng2019accurate} 
and shape-from-shading based~\cite{richardson2017learning,t2018extreme} methods, and remarkable results have been achieved. However, both methods could not apply on our task directly. Parametric methods suffer from the large diversity of geometry shapes in caricature cases. For SfS algorithms, the underlying physical model could not capture various painting styles of artists. 

\vspace{-4mm}
\paragraph{3D Caricature Generation}
Following the parametric based methods of normal face reconstruction, researchers further introduce deformation to enlarge the capability of representation~\cite{liu2009semi, wu2018alive, zhang2020landmark}. In \cite{liu2009semi}, a semi-supervised manifold regularization method is proposed to learn a regressive model for mapping from 2D real faces to the enlarged training set with 3D caricatures. Wu {\em et al.}~\cite{wu2018alive} formulate the 3D caricature generation as a problem of deformation from the standard 3D face. By introducing local deformation gradients, they build an intrinsic deformation representation with the capability of extrapolation. With the deformation representation, they construct an optimization framework to create caricature model guided by the landmark constraint. Following \cite{wu2018alive}, Zhang {\em et al.}~ \cite{zhang2020landmark} employ CNN to learn the deformation parameters of the intrinsic deformation representations. However, due to the lack of 3D caricature data, their works are still far from satisfaction. 

\vspace{-4mm}
\paragraph{3D Face datasets}
3D face datasets are of great value in face reconstruction tasks. In general, they could be categorized into synthetic and real captured datasets. For normal face, existing 3D datasets, including FaceWareHouse~\cite{cao2013facewarehouse} and Facescape~\cite{yang2020facescape}, are built from scanned 3D data, hence widely used in normal face tasks. They focus on the high accuracy and photo reality of the meshes. However, they could not be applied directly on caricature reconstruction. Researchers~\cite{wu2018alive,zhang2020landmark} tried to perform deformation on real 3D face models to construct synthetic exaggerated data. Although some reasonable results are achieved, they still suffer from the lack of diversity. To tackle this problem, we propose \dname{}, which is the first 3D caricature dataset built by artists, composed pairs of caricature images and meshes. Based on the dataset, 3D caricature shape could be learned in a model-free manner. We further propose a baseline method to reconstruct 3D mesh with uniform topology from single caricature image.

	\begin{figure}[t]
    \centering
    \includegraphics[width=8.3cm]{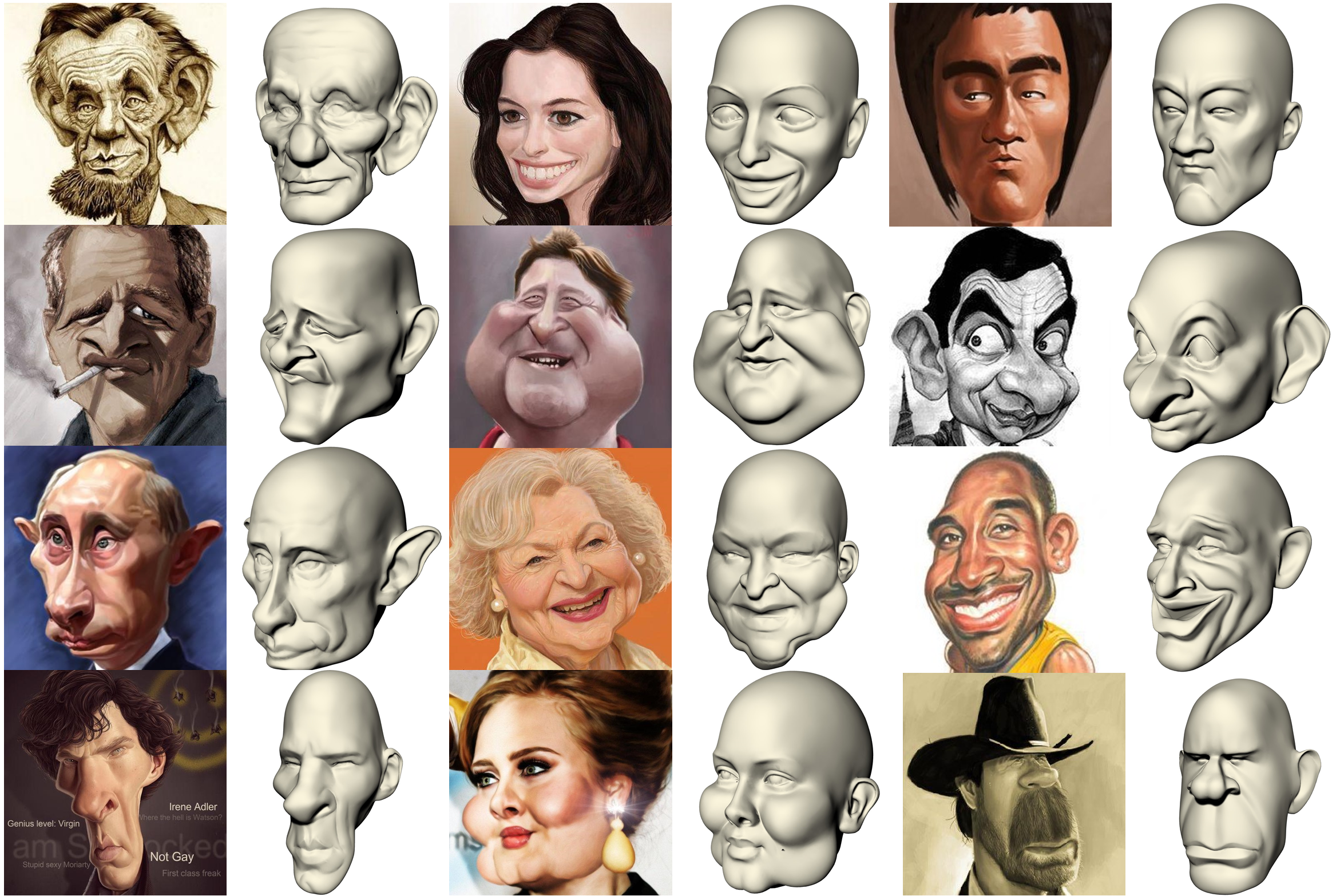}\\
    \caption{Sample caricature images and crafted 3D meshes in \dname{}. Images with diverse identities, geometry, and textural styles are collected. }
    \label{fig:sample}
    \vspace{-3mm}
\end{figure}

\section{Dataset}
We construct a dataset which contains 2,000 image-model pairs in total. All of the 3D models are annotated with 3D facial landmarks and poses w.r.t images. More details are introduced in the following aspects. 

\vspace{-4mm}
\paragraph{3D Model Collection} WebCaricature ~\cite{huo2017webcaricature} is the largest-to-date dataset of 2D caricatures. It contains around 6,000 caricature images with diverse identities, geometry, and textural styles. We first selected 2,000 images from them, further making them as diverse as possible. Then we recruited 4 paid expert Zbrush artists to create models according to images. The modeling is required to be matched with the image as much as possible, in projection manner. The contour lines for matching include edges of silhouette, lips, eyes, nose bottom and ears. It takes around 40 minutes for each model on average, and around 40 days are cost in total. Several image-model pairs sampled from our dataset are shown in Fig.~\ref{fig:sample}. Each model consists of 300,000~$\sim$~700,000 vertices. More examples can be found in the supplemental material.  

\begin{figure}
    \begin{center}
    \includegraphics[width=2.5in]{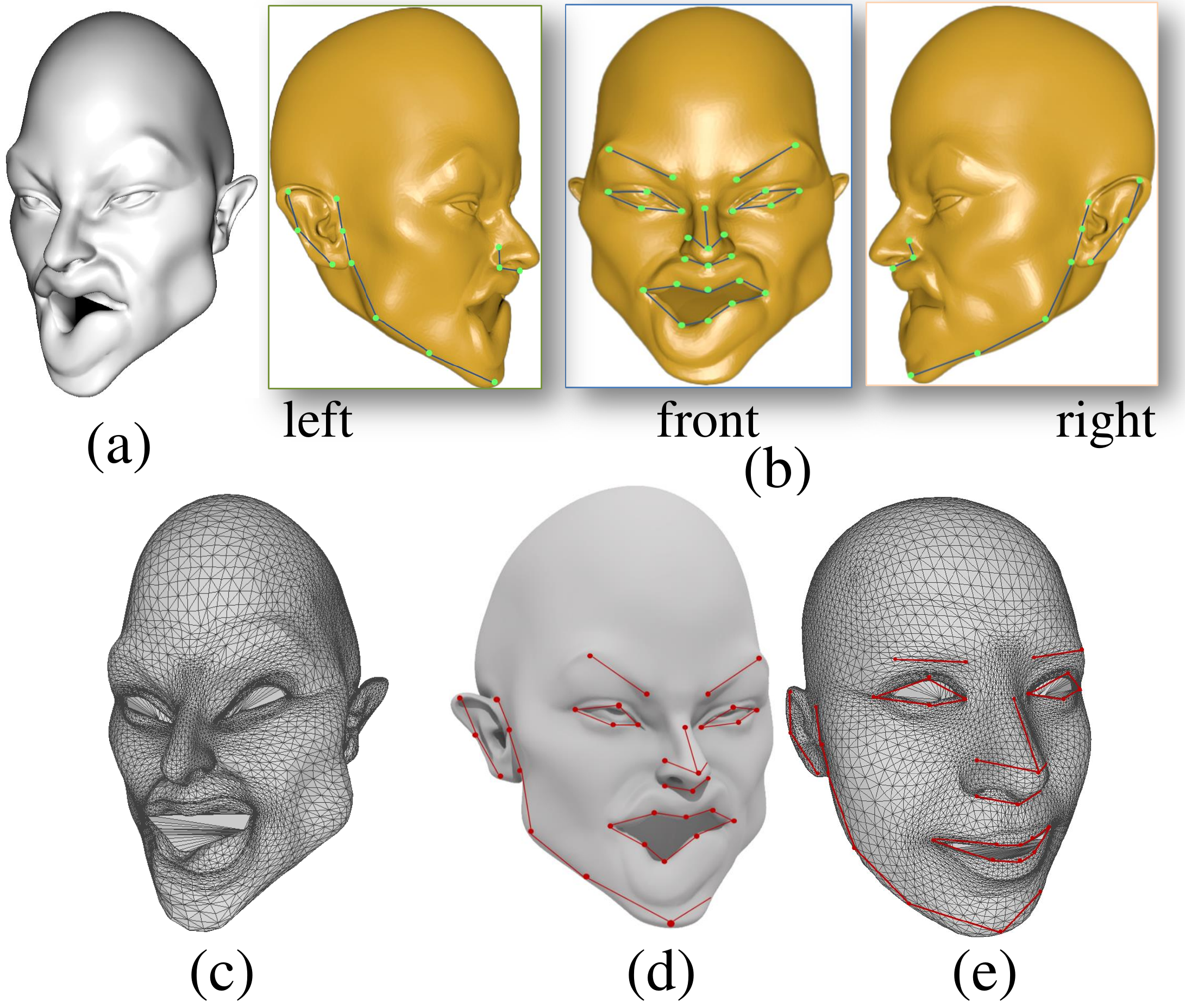}
  \caption{The process of 3D landmark annotation. A raw mesh (a) is rendered from the front, left and right view. Then 2D landmarks (b) are manually annotated for each view image. 3D landmarks are obtained by projecting the mesh into a specific view and searching for the closest point on surface (d). Guiding by corresponding 3D landmarks, the template mesh (e) is deformed into the shape of the raw mesh (d), to generate mesh with correct topology (c).}
  \label{fig:unification}
  \vspace{-6mm}
  \end{center}
\end{figure}

\vspace{-4mm}
\paragraph{Meshing Unification} To support building parametric space for our 3D caricature dataset, we unify the mesh topology for all models in two steps: 1) We first manually annotate 44 3D landmarks (see details in Fig.~\ref{fig:unification}) for each model; 2) The method of Non-rigid ICP~\cite{amberg2007optimal,dai2020statistical} is applied to register a pre-defined template mesh to each model, guided by the 3D landmarks. Due to the inherent difficulty to specify vertices on a 3D mesh, the landmark annotation is performed on 3 rendered views of the 3D shape. As described in \cite{booth20163d}, these 2D landmarks can be easily transformed into their corresponding 3D positions. The template mesh we use is from FaceWareHouse ~\cite{cao2013facewarehouse} that consists of 11,551 vertices. This procedure is illustrated by an example in Fig.~\ref{fig:unification}.


\vspace{-4mm}
\paragraph{Pose Annotation} 3D pose estimation from a single image is the premise of our reconstruction method (see Sec.~\ref{sec:basemethod}). It usually requires pose information supervision of the 3D face w.r.t the image. \dname{} also provides accurate pose labels for each 3D face mesh that annotated by artists manually. 

\vspace{-4mm}
\paragraph{Analysis of the dataset}
We quantitatively analyze our dataset by comparing the shape variations with two normal face datasets (FaceWarehouse (FWH) and FaceScape), as well as one synthetic caricature dataset, FaceWarehouse with deformation (Aug. FWH). 
We measure the shape variation using global and part variance. 
In particular, the variance is computed between the models and their corresponding mean shape of each dataset in terms of per-vertex displacement.
The results are presented below. 
The shape diversity of our dataset is richer than the normal ones. For most of the face regions, 3DCaricShop has larger shape variance than Aug. FWH. We will include this analysis in the revision.
\begin{table}[h]
\scriptsize
\vspace{-2mm}
\begin{center}
\begin{tabular}{|l|c|ccccc|c|}
\hline
Dataset & Global & Eye & Nose & Mouth & Ear & Cheek & Face\\
\hline\hline
FWH & 3.41 & 0.71 & 0.61 & 2.60 & 4.41 & 1.43 & 3.40\\
FaceScape & 2.17 & 0.36 & 0.15 & 2.63 & 5.57 & 1.24 & 2.27\\
\hline
Aug. FWH  & 5.06 & 1.98 & \textbf{6.29} & 2.07 & \textbf{9.38} & 5.26 & 5.10\\
\textbf{Ours} & \textbf{8.26} & \textbf{4.68} & 3.04& \textbf{10.90} & 9.02 & \textbf{8.27} & \textbf{6.95}\\
\hline
\end{tabular}
\end{center}
\vspace{-4mm}
\caption{Shape variance comparisons between different datasets. }
\label{tab:dataset}
\vspace{-4mm}
\end{table}







	\section{Methodology}
\label{sec:method}
\paragraph{Overview}

\label{sec:overview}
In this section, we introduce the proposed baseline method. Given an input caricature image $\mathbf{I}$, the task is to generate the corresponding 3D mesh $\mathbf{M}$. With the topologically uniform 3D meshes in \dname{}, a straight forward way to tackle the task is to  construct a PCA basis using the 3D Morphable Model algorithm \cite{blanz1999morphable} to build the caricature face space. However, such a space could not handle the large variation in our data. To capture the diversity of geometry in caricature, we employ Pixel-aligned Implicit Function (PIFu) \cite{saito2019pifu} to generate the 3D shape $\meshPIFu$ from the input image. Although the implicit function models the variation in targets, it could not ensure a uniform topology for the predictions. To achieve that, we register a template mesh $\meshTemp$ to $\meshPIFu$ using non-rigid registeration (NICP) \cite{amberg2007optimal}. Then the output of NICP is projected onto the pre-constructed PCA space, to alleviate deformation artifacts, such as self-intersections. We denote the output of NICP as $\meshNICP$ and that of PCA as $\meshPCA$. Considering the large difference between the template and target meshes, a sparse 3D landmark is needed in the stage of NICP. We propose a novel view-collaborative graph convolution network (VC-GCN) to predict key points $\mathbf{k}$ from the implicit mesh, where $\mathbf{k} \in \mathbb{R}^{44}$.


\begin{figure*}[htb]
  \begin{center}
      \includegraphics[width=\linewidth]{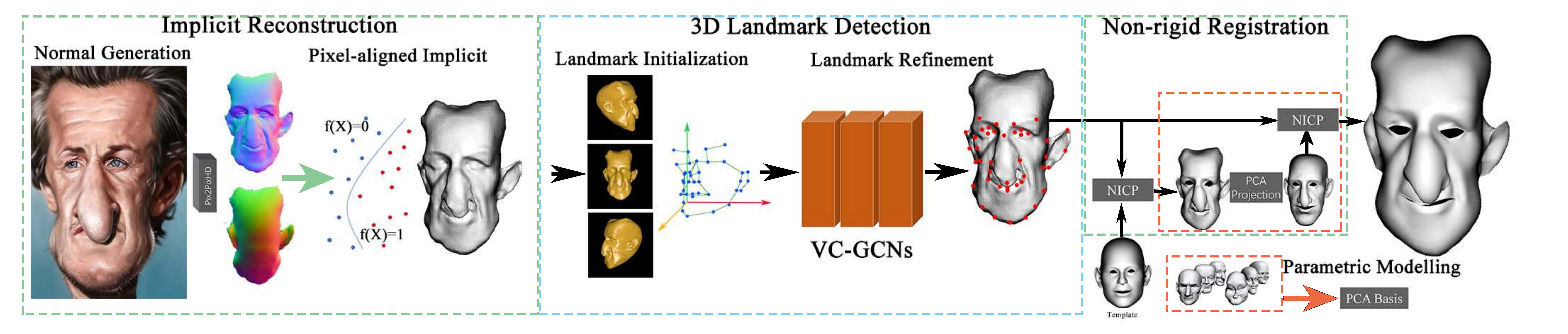}

  \end{center}
  \vspace{-6mm}
  \caption{ Pipeline of our framework. A detail-rich mesh is generated by implicit reconstruction from 2D caricature and the corresponding normal maps. Based on this mesh, 3D landmarks of the mesh are detected to guide the non-rigid registration, which deforms a template mesh to the target one. 
  }
  \label{fig:pipeline}
  \vspace{-4mm}
\end{figure*}

\subsection{The Baseline Approach}
\label{sec:basemethod}

\paragraph{Parametric Modeling}
Our parametric model space is built with standard 3D Morphable Model (3DMM) \cite{blanz1999morphable} algorithm.
Given $p$ caricature models with uniform topology and $N$ vertices on each mesh, principal component analysis (PCA) is performed on the shape matrix $\mathbf{S}_M \in \mathbb{R}^{3N\times p} $, which is formed by stacking the 3D coordinates of the $N\times p$ vertices. The generated $d$ eigen-vectors are employed as the shape basis $\mathbf{S}_i, i=1,2,...,d$, where $d$ is a hyper-parameter. The mean vector $\mathbf{\overline{S}}$ represents the mean shape in the mesh set. With this 3DMM, a novel caricature model $\mathbf{S}_N$ could be represented as follows:
$
    \mathbf{S_N}=\mathbf{\overline{S}}+\sum_{i=1}^{d}a_i\mathbf{S}_i,
$
where $\boldsymbol a = [a_1 \cdots a_d]^T$ is the vector of shape coefficients. 

\vspace{-3mm}
\paragraph{Implicit Reconstruction}
To capture the diversity of geometric variation in the 3D data, we adopt Pixel-aligned Implicit Function (PIFu) \cite{saito2019pifu} to reconstruct the underlying 3D shape from images. 
PIFu performs 3D reconstruction by estimating the occupancy of a dense 3D shape, which determines whether a point in 3D space is inside the model or not. Given a RGB image $\mathbf{I}$, its normal maps from the front view $\mathbf{F}$ and back view $\mathbf{B}$ are generated to strengthen the local details, by using a pixel2pixel-hd network \cite{wang2018high}. Then the implicit binary function $f(\mathbf{X},\mathbf{I},\mathbf{F},\mathbf{B})$ could be written as:
\begin{equation}
    f(\mathbf{X},\mathbf{I},\mathbf{F},\mathbf{B})=\begin{cases}
    1 ,& \text{if } \mathbf{X} \text{ is inside the mesh surface,}\\
    0, &\text{otherwise.}
    \end{cases}
\end{equation}
where $\mathbf{X}$ is a given 3D location in the continuous camera space. This function is modeled by a neural network.  
The loss function for training is formulated as:
\begin{equation}
\vspace{-1mm}
\mathcal{L}=\frac{1}{n}\sum^{n}_{i=1}|f(\mathbf{X}_i, \mathbf{I}_i,\mathbf{F}_i, \mathbf{B}_i)-f^{*}(\mathbf{X}_i)|^{2},
    \vspace{-0mm}
\end{equation}
where $f^{*}(\mathbf{X}_i)$ is the ground-truth occupancy.

\vspace{-2mm}
\paragraph{3D Landmark Detection for Registration}
The output meshes $\meshPIFu$ of the implicit function are not topologically uniform. In order to unify the topology, we adopt non-rigid registration\cite{amberg2007optimal} to deform a template $\meshTemp$ into the shape of $\meshPIFu$. As shown in \cite{amberg2007optimal}, without landmarks the cost function of registration could run into a local minimum, where the template is collapsed onto a point on the target surface. Thus it is important to introduce the 3D landmarks of both meshes to guide the deformation. We design a novel framework to detect 3D landmarks for $\meshPIFu$. In short, we propose to perform detection on the rendered views of $\meshPIFu$ to leverage the effectiveness of image-based CNN techniques. The process would be detailed in \ref{sec:detection}.

\vspace{-3mm}
\paragraph{Landmark-guided Registration}
 Since the huge difference between $\meshTemp$ and $\meshPIFu$, the deformation is likely to generate artifacts on meshes, such as self-intersection. To resolve this problem, we iteratively perform NICP and PCA projection of the results to obtain $\meshNICP$ and $\meshPCA$. After projection, we obtain a deformed template which is closer to $\meshPIFu$ in shape. Fig. \ref{fig:pipeline} illustrates the process of the progressive deformation.

\subsection{View-collaborative 3D Landmark Detection}
\label{sec:detection}

\begin{figure*}
\label{fig:lmdetect}
 \begin{center}
      \includegraphics[width=0.95\textwidth]{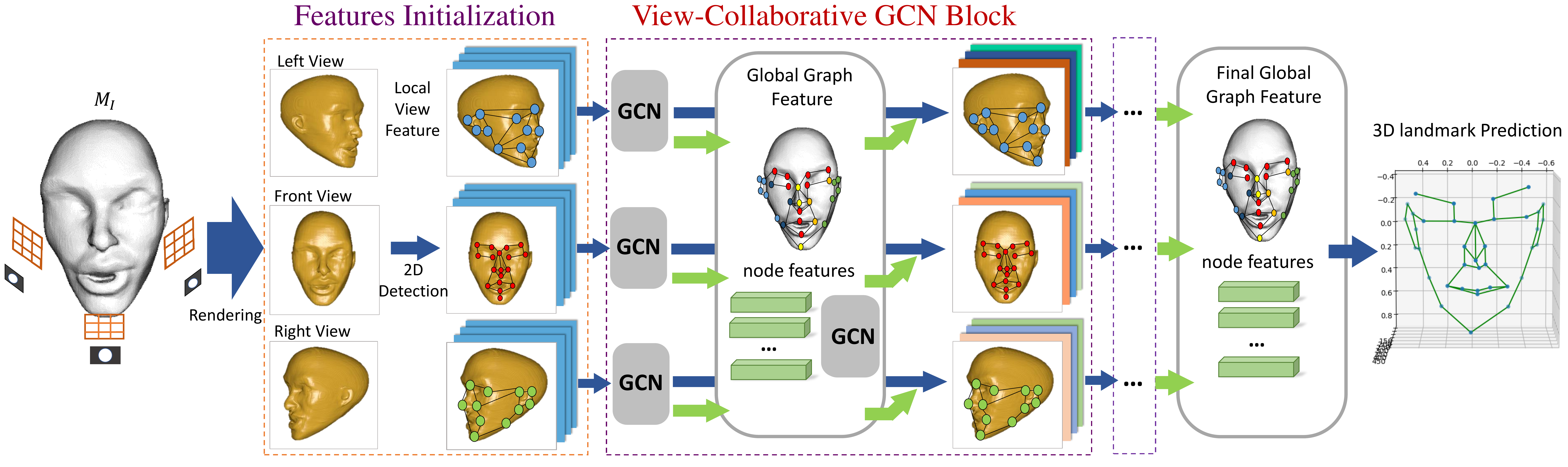}
  \end{center}
  \vspace{-5mm}
  \caption{The pipeline of View-collaborative 3D landmark Detection. The initial 3D landmaks are obtained by exploiting the predicted 2D landmarks from 3 rendering views. A novel cascaded VC-GCN blocks is used to fuse the features form each view and sends the aggregated features into a GCN head layer for 3D displacement decoding.}
  \label{fig:vcgcn}
  \vspace{-3mm}
\end{figure*}

In this section, we discuss more details about how to detect 3D landmarks from $\meshPIFu$, which is the key to supporting the procedure of landmark-guided registration. A straightforward way for this detection is directly applying point-based CNN (e.g., SparseConv~\cite{liu2015sparse}) to estimate landmark-ware heatmap on mesh vertices. However, due to the inherent difficulty to conduct CNN on a mesh, this approach tends to produce inaccurate results. We thus propose to perform detection on the rendered views of $\meshPIFu$ to leverage the effectiveness of image-based CNN techniques. Coarse locations of the 3D landmarks can be obtained from detected 2D landmarks on those views. More importantly, a stack of View-Collaborative GCN block (VC-GCN) is novelly designed to aggregate and enhance information from multiple views for accurate 3D landmarks locations. As illustrated in Fig. \ref{fig:vcgcn}, single view graph features (local features) are first extracted for initialization. Then, these local features are enhanced in a progressive manner by continually 
fusing global information into each view. The final local features are aggregated into the global graph feature for 3D landmark prediction.      

\vspace{-5mm}
\paragraph{Initialization Stage}
In this part, more details about local feature initialization are introduced. Given 2D images rendered from 3 views $ \{\text{front, left, right} \}$ ($\{ f,l,r \}$ for simplicity), we first utilize a 2D landmark detector \cite{guo2019pfld} to estimate the 2D landmarks $\mathbf{\hat{P}}^v \in \mathbb{R}^{k^v \times 2}$, where $k^v$ denotes the key point number under the view $v \in \{ f,l,r \}$. Next, 3d landmarks on the mesh are located using the projection matrix of each view. We use the above landmarks which exist in all local views to build local graphs. After that, to enrich the information of each graph node, we extract features from the feature maps generated by the landmark detector for each node, according to their 2D coordinates. Eventually, the initial local view features $\mathbf{F}_{init}^v \in \mathbb{R}^{k^v \times (C+3)}$ for VC-GCN can be produced by concatenating the 3D landmark locations $\mathbf{L}^v$ with related node features under each view $v$, where $C$ is the image feature dimension.



\vspace{-2mm}
\paragraph{View-Collaborative GCN}
In order to provide global information for each view, we aggregate 3 local features into a global graph feature. Then the global feature is fused into each view to enhance the local view feature. This procedure is performed by a View-Collaborative GCN block.  
In each VC-GCN block, local features are first sent into several GCN layers for better representations. 
The layer-wise operation in GCN is defined as:
\begin{equation}
    \mathbf{F}^v_{\ell+1} = \sigma\left({\mathbf{D}^v}^{-\frac{1}{2}} \mathbf{A}^v {\mathbf{D}^v}^{-\frac{1}{2}} \mathbf{F}^{v}_{\ell} \mathbf{W}^{v}_{\ell}\right)
\label{equ:gcn_layer}
\end{equation}
where ${\mathbf{A}^v}$ is the adjacency matrix with self-loops, ${\mathbf{D}^v}$ is its diagonal node degree matrix to normalize $\mathbf{A}^v$, ${\mathbf{F}^v_{\ell}}$ represents the local feature in layer $\ell \in \{0,1,\dots,\ell'\}$ under the view $v$, ${\mathbf{W}_\ell}$ is a trainable parameter matrix for linear projection, and $\sigma(\cdot)$ represents the non-linear activation operation. 
Then the obtained local features are combined into a global graph feature. For each node in the 3D landmark, its feature can be drawn from the local feature of the corresponding node under one view. Note that for the node that shared in different views, its feature is set as the average of multiple local features. The combined global graph feature is then strengthened through several GCN layers, with same operations as in Equation \ref{equ:gcn_layer}. 
Hence, the process of feature aggregation now can be formulated as following:
\begin{equation}
    \begin{split}
        \mathbf{F}_{\ell'}^{v}&=\text{GCN}^v(\mathbf{F}_{0}^{v}), v = f,l,r;\\
        \mathbf{F}_{\ell'}^{g}&=\text{GCN}^g(f_\text{comb}(\mathbf{F}_{\ell'}^{f},\mathbf{F}_{\ell'}^{l},\mathbf{F}_{\ell'}^{r})). \\ 
    \end{split}
\end{equation}
where $f_\text{comb}(\cdot)$ denotes the combination operation, $\mathbf{F}_{\ell'}^{g}$ is the strengthened global features. Note that the input $\mathbf{F}_{0}^{v}$ is set as the initial local view feature $\mathbf{F}_{init}^v$ in the first VC-GCN block, and is set as the prior output features in subsequent blocks.

In the second step, strengthened global features are fused into local features of each view in a non-local manner \cite{wang2018non}, so that global information can guide the model to learn more representative local features. The enhanced local features $\mathbf{F}_\text{enh}^{v}$ of the view $v$ can be obtained as following, where the non-local fusion operation is denoted as $f_\text{n-loc}$:
\begin{equation}
    \begin{split}
        \mathbf{F}_\text{enh}^{v} = f_\text{n-loc}(\mathbf{F}_{\ell'}^{g}, \mathbf{F}_{\ell'}^{v}), v = f,l,r.\\ 
    \end{split}
\end{equation}
More details about the non-local fusion operation are described in the supplementary materials. 

It usually takes numerous glimpses to adjust key points to construct a 3D face, even for an expert artist. Thus, several VC-GCN blocks are stacked to progressively enhance local features. In the connection of two blocks, the enhanced local features $\mathbf{F}_\text{enh}^{v}$ from the former block are taken as the input $\mathbf{F}_{0}^{v}$ of the later block. 



\vspace{-2mm}
\paragraph{Loss Function}
Given the enhanced local features from the last VC-GCN block, we combine and strengthen them using GCN layers to obtain the final global graph features. Next, it is multiplied by a GCN head layer to get the 3D landmark estimation $\mathbf{\hat{L}}^g \in \mathbb{R}^{N\times 3}$. The predicted 3D landmarks are supervised by 3D and 2D landmark ground truth simultaneously, which leads to more accurate prediction. 
We now formulate the loss function for the view-collaborative 3D landmark detection training as following:

\vspace{-3mm}
\begin{equation}
    \begin{split}
        \mathcal{L}=\sum_{i\in \Omega, v}(&\mathcal{L}_\text{detect}(\mathbf{\hat{P}}_i^v,\mathbf{P}_i^v) \\
        +&\mathcal{L}_\text{3D}(\mathbf{\hat{L}}_i^g, \mathbf{L}_i^g)\\
        +&\mathcal{L}_\text{2D}(\mathbf{M}^v(\mathbf{\hat{L}}_i^{g}),\mathbf{P}_i^v)),
    \end{split}
\end{equation}
where $\Omega$ is the training set, $i$ is the subscript indicating each training sample, $\mathcal{L}_\text{detect}$ denotes the 2D landmark detection error in the initialization stage, $\mathcal{L}_\text{3D}$ and $\mathcal{L}_\text{2D}$ represents the 3D landmark prediction error in 2D and 3D space, respectively, and $\mathbf{M}^v$ is the projection matrix to obtain 2D landmarks from 3D landmarks under the view $v$. Note that all loss terms are in smooth-$l_1$ form.

	\section{Experiments}
\label{sec:experiments}
\paragraph{Implementation details}
The proposed framework is trained on our \dname{}. The dataset is separated into 1,600 for training and 400 for testing. The weights of $\mathcal{L}_{detect}$, $\mathcal{L}_{2D}$ and $\mathcal{L}_{3D}$ are set to $0.1$, $0.8$ and $1.0$ respectively. To train the network for learning the implicit reconstruction, a RMSProp optimizer is adopted with learning rate 0.001, and the network is pre-trained with the mini-batch size 2 for 80 epochs. During the training of 2D landmark detection network, an Adam optimizer is used. The learning rate is set to 0.0001 with a cosine decay, and the mini-batch size is set to 24 for 30 epochs. After that, the whole framework is trained in an end-to-end manner with the same strategy as above. 

\begin{figure*}
 \begin{center}
      \includegraphics[width=0.93\textwidth]{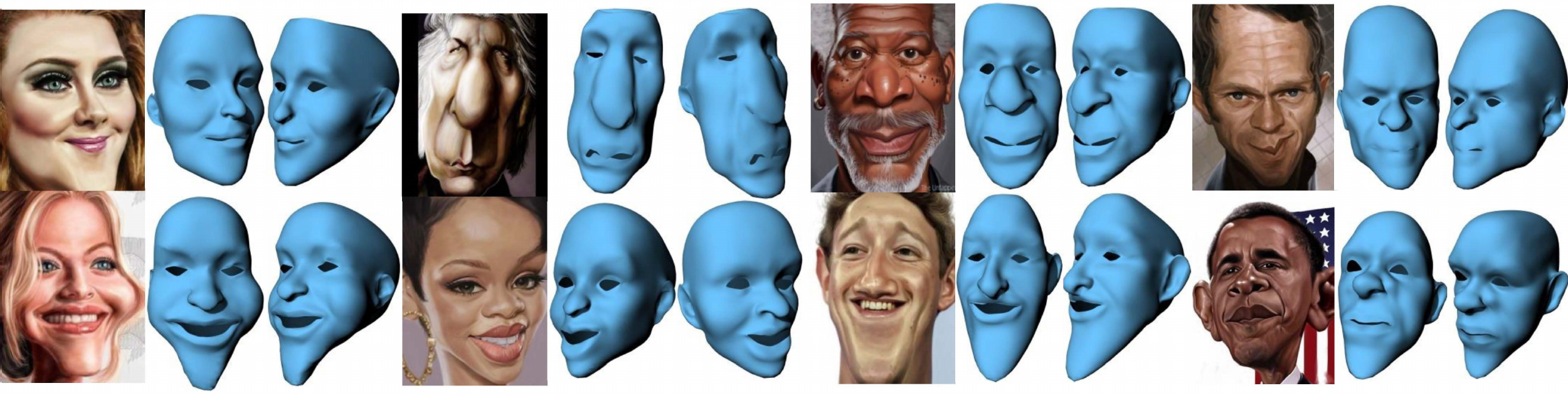}
  \end{center}
  \vspace{-4mm}
  \caption{Results gallery for our framework on \dname{}. The framework has the capability to reconstruct 3D shapes from caricature images with diverse texture and geometry shapes.}
  \label{fig:results_gallery}
  \vspace{-4mm}
\end{figure*}

\vspace{-2mm}
\paragraph{Results Gallery}
We present some typical results of the proposed framework in Fig.~\ref{fig:results_gallery}. As illustrated, our method is robust to caricature images with diverse textures. It can also recover diversified geometric features, such as the exaggerated nose in the second sample of the first row, and the sharp long chin in the third sample of the second row Fig.~\ref{fig:results_gallery}.

\subsection{Comparisons with the State-of-the-arts}
\label{sec:compare}
We qualitatively and quantitatively compare the results of our method with a variety of state-of-the-art 3D caricature reconstruction approaches on 3DCaricShop testing set, including linear parametric model (3DMM) \cite{t2018extreme, blanz1999morphable}, depth map (DF2Net) \cite{zeng2019df2net}, deformation representation (AliveCaric-DL) \cite{zhang2020landmark}, and implicit function (PIFu) \cite{saito2019pifu} based methods.

\begin{figure*}
 \vspace{-2mm}
 \begin{center}
      \includegraphics[width=0.95\textwidth]{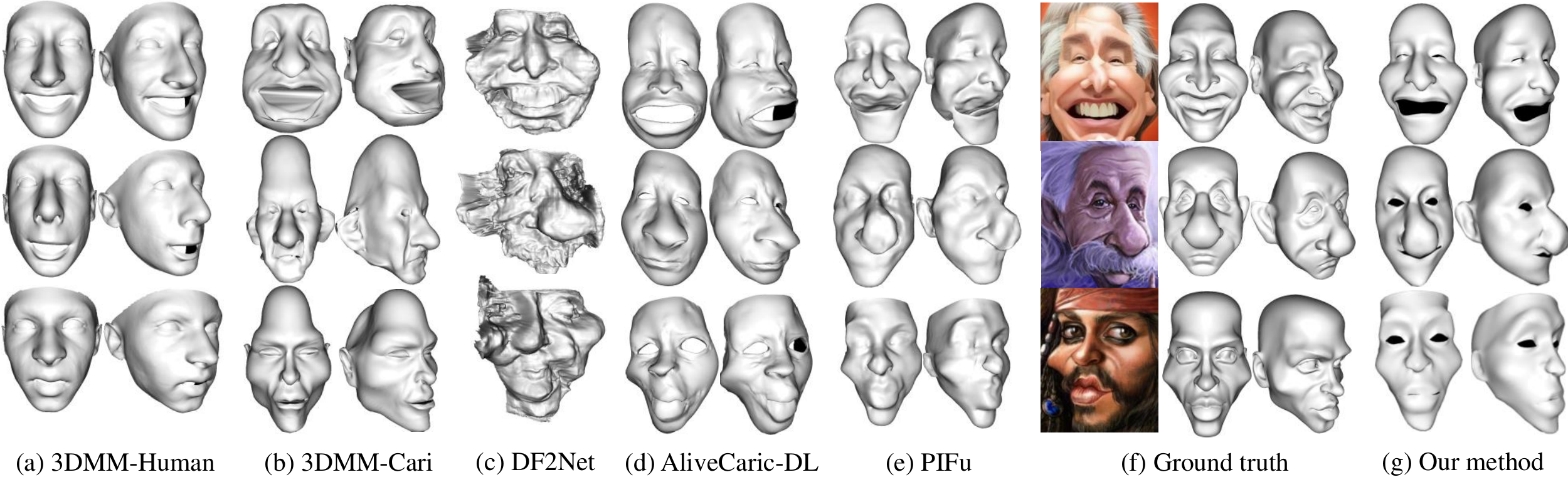}
  \end{center}
  \vspace{-2mm}
  \caption{Qualitative results of our method compared with state-of-the-art methods, including (a) 3DMM-Human \cite{t2018extreme}, (b) 3DMM-cari \cite{blanz1999morphable}, (c) DF2Net \cite{zeng2019df2net}, (d) AliveCaric-DL \cite{zhang2020landmark} and (e) PiFu \cite{saito2019pifu}, on \dname{}. By incorporating deep models with parametric space constraint, our method (g) can reconstruct highly exaggerated geometry without distinct artifacts.}
  \vspace{-3mm}
  \label{fig:comparison}
\end{figure*}

\vspace{-2mm}
\paragraph{Qualitative results}
In Fig.~\ref{fig:comparison}, we visualize some results of caricature reconstruction on images from 3DCaricShop. Among the parametric methods, the nonlinear deformation representation \cite{zhang2020landmark} based model outperforms the linear ones on fitting the exaggerated input images, but it is still not precise enough due to its limited expressiveness. Besides, other deep learning based approaches such as DF2Net and PIFu unavoidably yield artifacts like hollows and spikes. However, our method introduces the constraint of PCA parametric space into the deep model, thus can produce highly exaggerated local details upon the foundation of a plausible global shape.

\vspace{-2mm}
\paragraph{Quantitative Results}
Considering other methods for comparison only reconstruct the face area, 
we adopt average point-to-surface Euclidean distance (P2S) as the evaluation metric, which measures the unidirectional distance from the source set to the target set. The average point-to-surface Euclidean distance can be computed as:
\begin{equation}
    d_{P2S}(P, S)=\frac{1}{\Vert{P}\Vert}\sum_{p\in {P}}\min_{p'\in {S}}\Vert{p-p'}\Vert_2
\label{equ:chamfer dist}
\end{equation}
where $P$ is the vertex set of the reconstructed mesh and $S$ is the corresponding ground truth surface. Besides, due to the mismatch in orientation and scale between the generated meshes and ground truth, before calculation, Procrustes alignment is performed and scaling is estimated based on least square error.
\begin{table}
\begin{center}
\footnotesize
\begin{tabular}{|l|c||l|c|}
\hline
Methods & P2S & Methods & P2S\\
\hline\hline
$\text{3DMM}_\text{human}$ & 0.295 & $\text{PiFu}_\text{head}$ & 0.153  \\
$\text{3DMM}_\text{cari}$ & 0.104 & $\text{PiFu}_\text{face}$ & 0.126 \\
DF2Net & 0.273 & $\text{Ours}_\text{head}$ & 0.065   \\
AliveCaric-DL & 0.067 & $\textbf{Ours}_\textbf{face}$ & $\textbf{0.037}$\\
\hline
\end{tabular}
\end{center}
\vspace{-3mm}
\caption{Quantitative evaluation on 3DCaricShop. Note that the meshes generated by our method and PiFu contain the entire head area which are the same with ground truth, whereas the other methods in our experiment only recover the frontal face, thus we provide two versions of results(i.e. head and face) on our method and PiFu for a more solid comparison. The results are averaged on test data.}
\label{tab:eval}
\vspace{-4mm}
\end{table}
As shown in Table~\ref{tab:eval}, our method achieves the smallest P2S on the 3DCaricShop. 

\
\subsection{Ablation Studies}
In this section, we perform ablation studies on the proposed 3D landmark detection framework and landmark guided registration process. The results show the effectiveness and robustness of our pipeline.

\vspace{-2mm}
\paragraph{3D landmark detection}
We analyze five variants of our framework: 1) directly using the 3D landmarks selected from predicted 2D landmarks without subsequent refinement, denoted as `w/o GCN refinement'; 2) utilizing voxel-based method \cite{moon2018v2v} to estimate 3D heatmaps, denoted as `V2V'; 3) employing a global graph to refine the 3D landmarks from the first setting, without using VC-GCN block, which denoted as `Global only'; 4) Only using local index to gather local features from global view, rather than using non-local operations for local feature enhancement, denoted as `w/o G2L'; 5) The basic setting, denoted as Basic.

The metric we evaluate the results is mean per joint position  error (MPJPE) which is defined  as a Euclidean distance between predicted and ground truth 3D landmarks after root joint alignment. The root joint we define is the top of nose. This metric measures how accurately the root-relative 3D landmark estimation is performed.
The quantitative results are listed in Table. \ref{abla:landmark}. It confirm the effectiveness of each components in the design of our 3D landmark detector. 


\begin{figure}
    \begin{center}
        \includegraphics[width=3in]{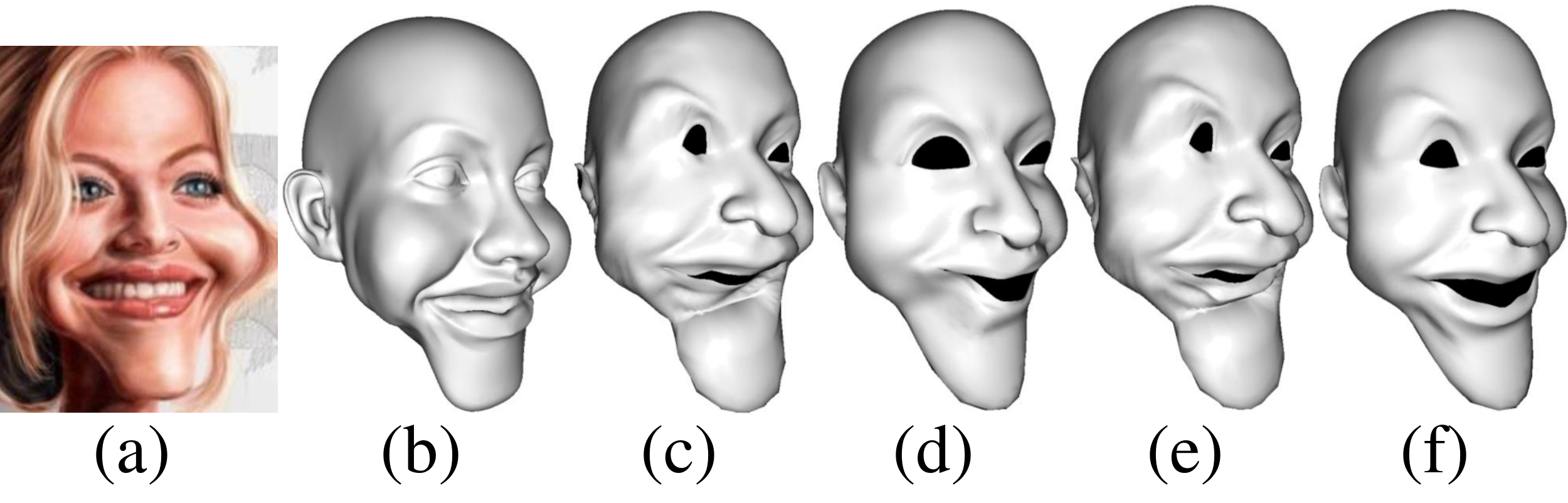}
    \end{center}
  \vspace{-4mm}
  \caption{Ablation experiments on 3D landmark detection: (a) input; (b) ground truth; (c) pure projection; (d) global only; (e) w/o G2L; (f) ours. Our method could capture the face geometry more accurately. }
  \label{fig:ablation_lm}
\end{figure}

\begin{figure}
    \begin{center}
        \includegraphics[width=3in]{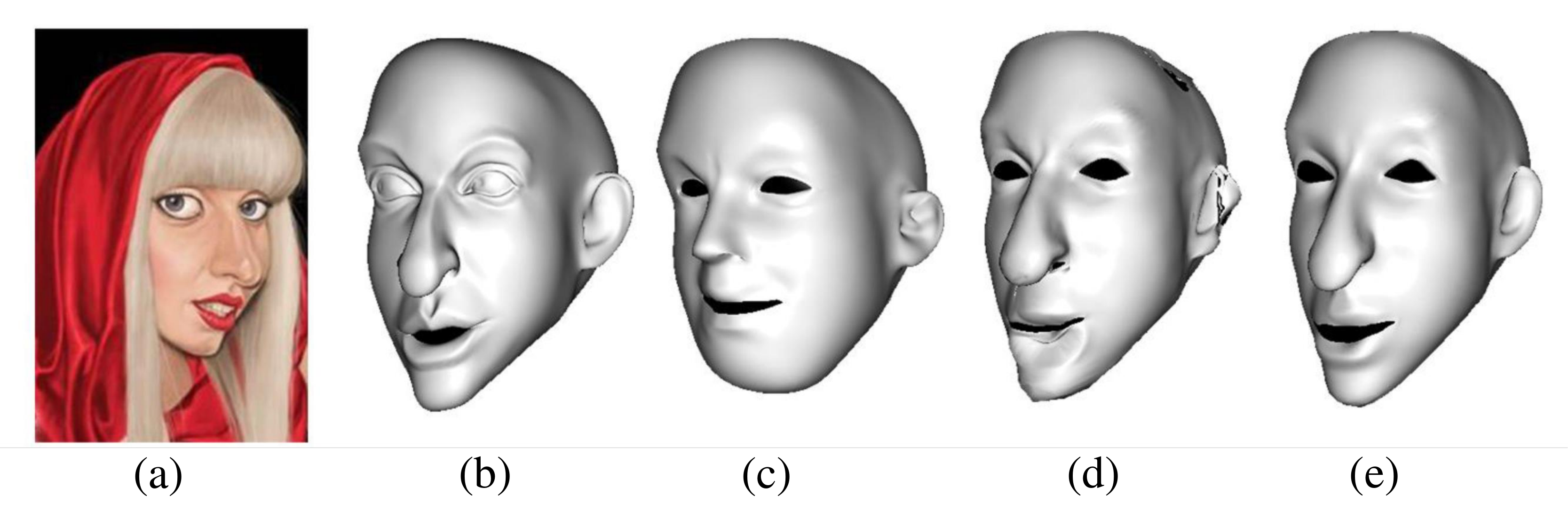}
    \end{center}
  \vspace{-4mm}
  \caption{Ablation experiments on registration: (a) input; (b) GT; (c) NICP w/o landmark; (d) NICP w/o PCA projection; (e) ours. A better result is obtained with reasonable topology (e.g., the nose) by using PCA projection.}
  \label{fig:ablation_reg}
  \vspace{-2mm}
\end{figure}

\vspace{-2mm}
\paragraph{On landmark guided registration}
We evaluate three kinds of registration process: 1) directly perform NICP without landmarks information; 2) perform landmark-guided NICP without PCA projection; 3) the process used in our method.
The visualized results are shown in Fig.~\ref{fig:ablation_reg}. As \cite{amberg2007optimal} suggested, without landmark information, NICP could not capture the large discrepancy between $\meshTemp$ and $\meshPIFu$. Besides, the deformation without PCA projection is likely to generate meshes with self-intersection. In contrast, our method could obtain meshes with higher quality, and capture enough shape information in $\meshPIFu$. For example, the artifacts on ears in  Fig.~\ref{fig:ablation_reg}(d) are eliminated, while the shape of nose is more consistent with both the groud truth mesh and the input caricature image.
\begin{table}
\footnotesize
\begin{center}
\begin{tabular}{|l|c||l|c|}
\hline
Methods & MPJPE & Methods & MPJPE\\
\hline\hline
w/o GCN Refinement & 0.451 & Global only & 0.373\\
V2V \cite{moon2018v2v} & 0.407 & w/o G2L & 0.358\\
\hline
& & \textbf{Basic} & \textbf{0.291} \\
\hline
\end{tabular}
\end{center}
\vspace{-4mm}
\caption{Ablation study for 3D landmark detection.}
\label{abla:landmark}
\vspace{-2mm}
\end{table}

\begin{table}
\begin{center}
\footnotesize
\begin{tabular}{|l|c|}
\hline
Methods & P2S \\
\hline\hline
Ours w/o Landmark & 0.074\\
Ours w/o PCA projection & 0.076\\
\hline
\textbf{Ours} & \textbf{0.065}\\
\hline
\end{tabular}
\end{center}
\vspace{-2mm}
\caption{Ablation study for landmark-guided registration. The similar data implies the improvement of our method concentrate on the detail structures. }
\label{abla:registration}
\vspace{-4mm}
\end{table}

\section{Conclusions}

We construct a new dataset and benchmark, called \dname{}, for single-view 3D reconstruction from caricature images. 
\dname{} is the largest collection by far of 3D caricature models crafted by professional artists.
It consists of 2,000 high-quality and diversified 3D caricatures that are richly labeled with paired 2D caricature image, camera parameters, and 3D facial landmarks. 
A novel baseline approach is also presented to validate the usefulness of the proposed dataset.
It combines the merits of flexible implicit functions and the robust parametric mesh representation.
Specifically, we transfer the details from implicit reconstruction to a template mesh with the help of VC-GCN that accurately predicts 3D landmarks for the implicit mesh.
Extensive benchmarking on our dataset has been performed including a variety of popular approaches.
We found that reconstructing 3D caricature from a single 2D caricature image is a highly challenging task with ample opportunity for improvement.
We hope \dname{} and our baseline approach could shred light on future research in this field.
	\section{Acknowledge}
	This work was supported in part by the National Key R\&D Program of China with grant No. 2018YFB1800800, the Key Area R\&D Program of Guangdong Province with grant No. 2018B030338001, Guangdong Research Project No. 2017ZT07X152, the National Natural Science Foundation of China 61902334, Shenzhen Fundamental Research (General Project) JCYJ20190814112007258.

	
	{\small
		\bibliographystyle{ieee_fullname}
		\bibliography{paper}
	}
		
    \section*{Appendix}
\appendix
\section{Structure of G2L Network}
\label{sec:VCGCN}
\begin{figure}
    \begin{center}
    \includegraphics[width=2.6in]{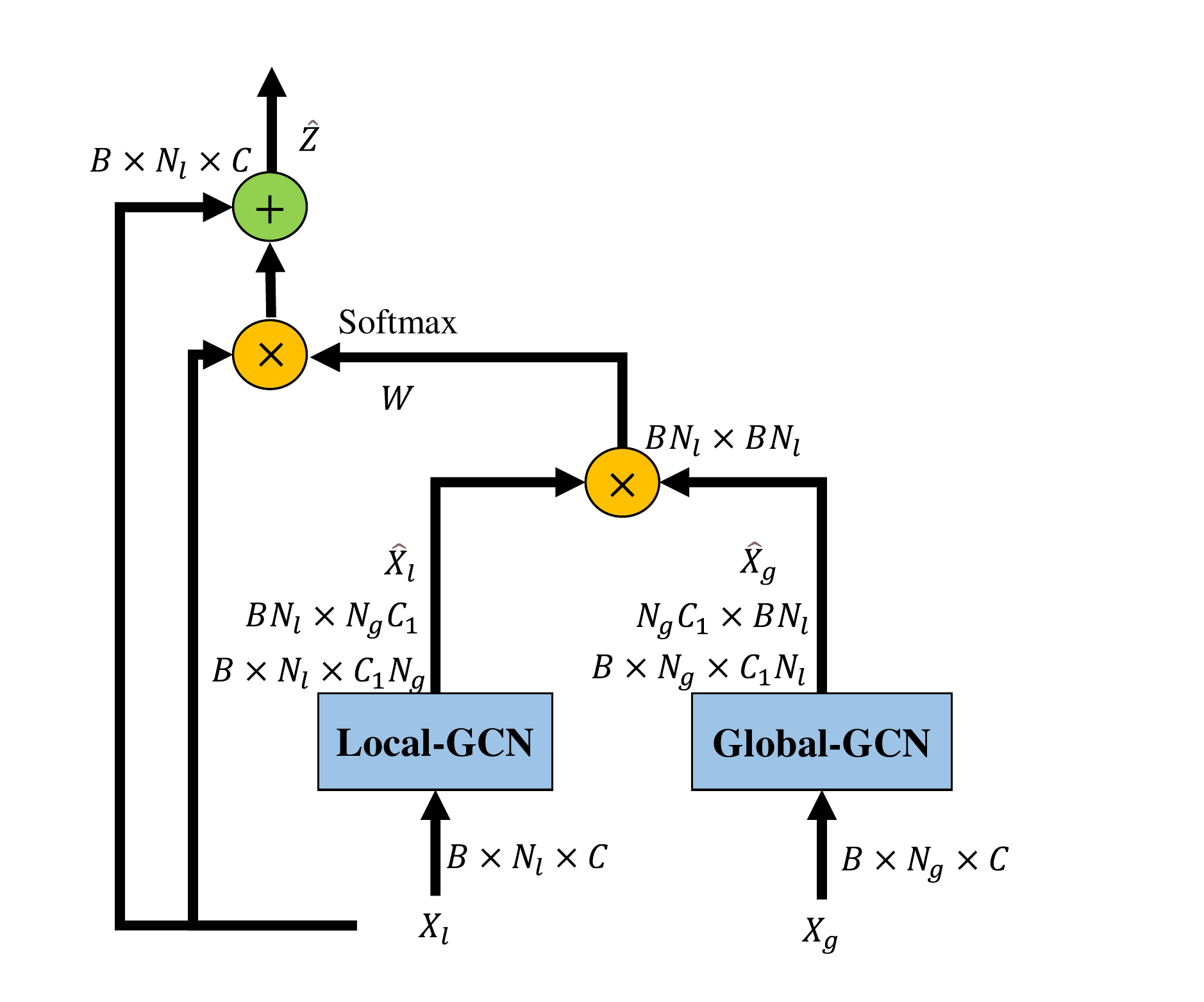}
    \vspace{-4mm}
  \caption{The pipeline of Global to Local(G2L) module in VC-GCN. $\mathbf{X}_l$ means local-view features and ${\mathbf{X}_g}$ represents global-view features. $\bigotimes$ is matrix multiplication and $\bigoplus$ stands for the pixel-wise addition. The trainable global-to-local weights are defined as W while $\mathbf{\hat{X}}_l$ and $\mathbf{\hat{X}}_g$ represents the outputs of local-GCN and Global-GCN respectively. $\hat{Z}$ is the updated local features fused with global ones. B is the batch size of inputs. $N_l$ and $N_g$ represent the number of nodes of local and global graph, with C defined as the number of feature channels.}
  \label{fig:supple_g2l}
  \vspace{-4mm}
  \end{center}
\end{figure}

In this section, we illustrate the structure of the Global to Local (G2L) moduleof VC-GCN. As shown in Fig. \ref{fig:supple_g2l}, the outputs of local-view GCN $\mathbf{X}_l$ and those of global-view GCN  $\mathbf{X}_g$ are fed into G2L network. 
First, we change the channels of global and local features ($\mathbf{\hat{X}}_g$ and ${\mathbf{\hat{X}}}_l$ respectively) by local-GCN and global-GCN, 
of which the structures are the same as that employed in the whole pipeline. Then, trainable G2L weights $\mathbf{W}$ are obtained by matrix multiplication between $\mathbf{\hat{X}}_g$ and ${\mathbf{\hat{X}}}_l$ , followed by a softmax operation. Finally, we get the updated local-view features $\mathbf{\hat{Z}}_l$ processed by the following formula:

\begin{equation}
\vspace{-2mm}
    \mathbf{\hat{Z}}_l=\mathbf{W}\bigotimes \mathbf{X}_l+\mathbf{X}_l,
    \vspace{-1mm}
\end{equation}

\noindent where $\bigotimes$ means matrix multiplication. In Fig. \ref{fig:supple_g2l}, $\mathbf{B}$ is the batch size of inputs. $N_l$ and $N_g$ represent the number of nodes in the local and global graph, with $C$ and $C_1$ defined as the number of feature channels. Empirically, $C_1$ is set to 32, considering the trade-off between efficiency and accuracy. 

\vspace{-2mm}

\section{More Qualitative Results}
\begin{figure}
    \begin{center}
    \includegraphics[width=3.2in]{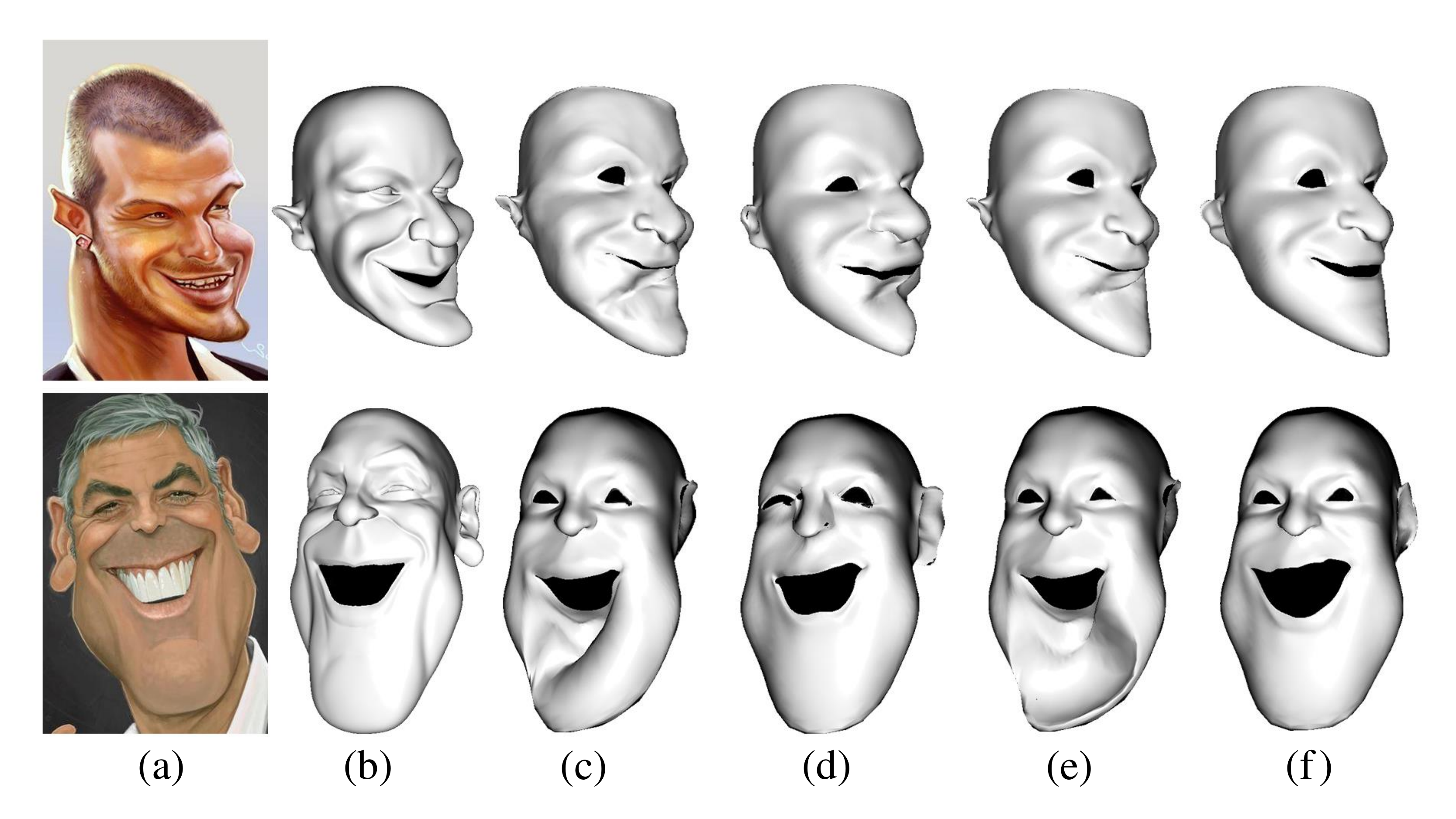}
    \end{center}
  \caption{ Visualized reconstruction results using different setting of 3D landmark detection, with (a) input; (b) GT; (c) pure projection; (d) global only; (e) w/o G2L; (f) ours. It demonstrates that our method could capture the face geometry more accurately. }
  \label{fig:supp_ablation_lm}
\end{figure}

\begin{figure}
    \begin{center}
        \includegraphics[width=3.2in]{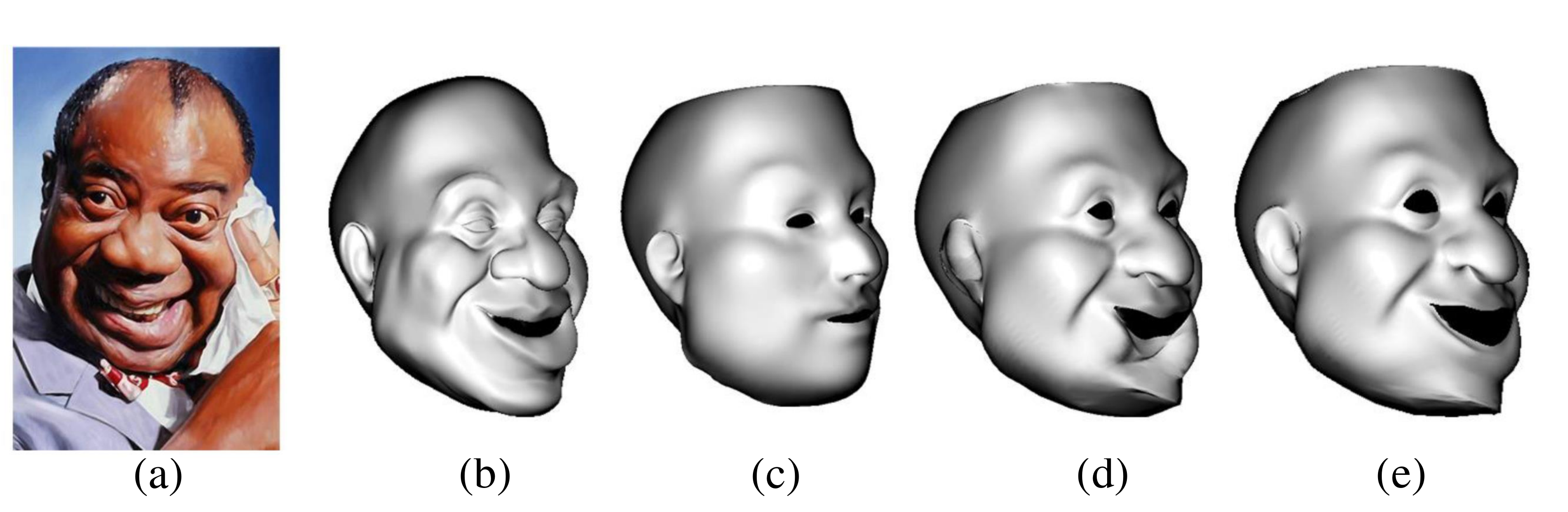}
    \end{center}
  \caption{Visualized reconstruction results on different setting of registration: (a) input; (b) GT; (c) NICP w/o landmark; (d) NICP w/o PCA projection; (e) ours. It demonstrates that a better template is obtained with finer details (e.g., the nose) by using PCA projection.}
  \label{fig:supp_ablation_rg}
\end{figure}

 Fig.~\ref{fig:supp_ablation_lm} shows that the reconstruction results using our proposed 3D landmark localization approach could capture the large exaggerations more accurately than other settings. For example, the long chin of the second sample is not distorted.
 Fig.~\ref{fig:supp_ablation_rg} shows the necessity of landmark-guided registration. Without 3D landmarks, the outputs of NICP \cite{amberg2007optimal} fail to fit the accurate shape of the face, and PCA \cite{blanz1999morphable} projection helps to further reduce artifacts in the final results. 

\begin{figure}
    \begin{center}
    \includegraphics[width=3.2in]{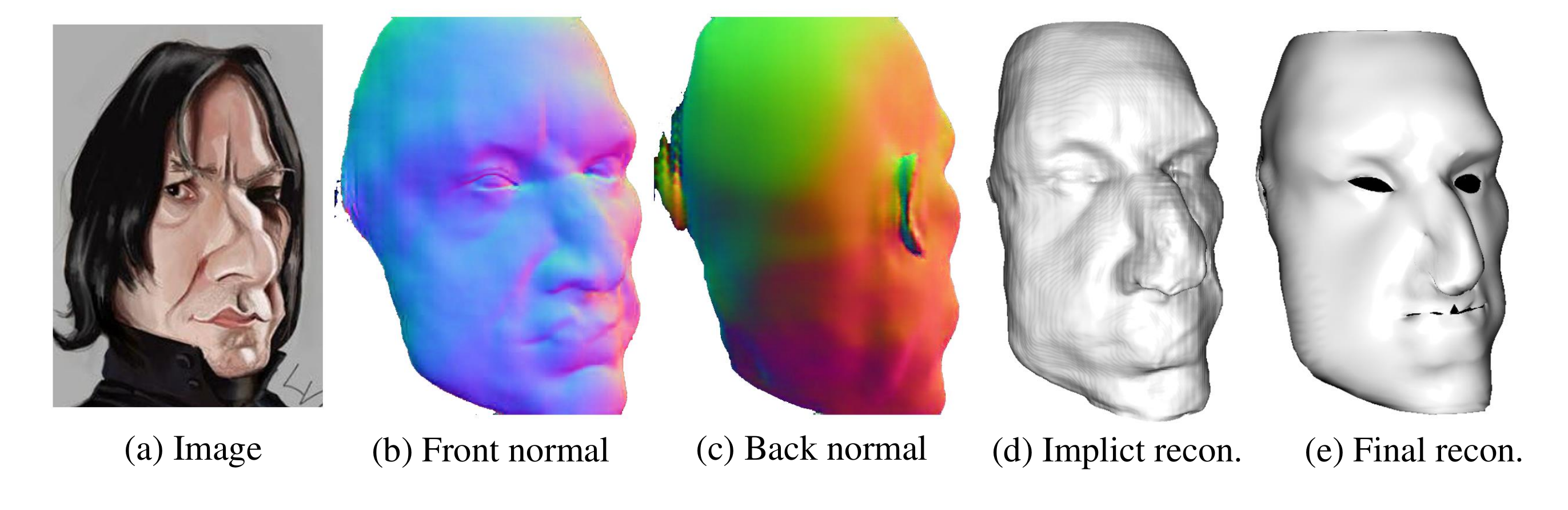}
  \caption{A Failure case. The normal is broken on the ear and blur on the mouth, generating a low quality  reconstruction.
  }
  \label{fig:failed}
  \end{center}
\end{figure}

We show a failure case in Fig.~\ref{fig:failed} where the estimated normal map is blurry, especially at the mouth region, that leads to inaccurate result.


\begin{figure*}
 \begin{center}
      \includegraphics[width=0.90\textwidth]{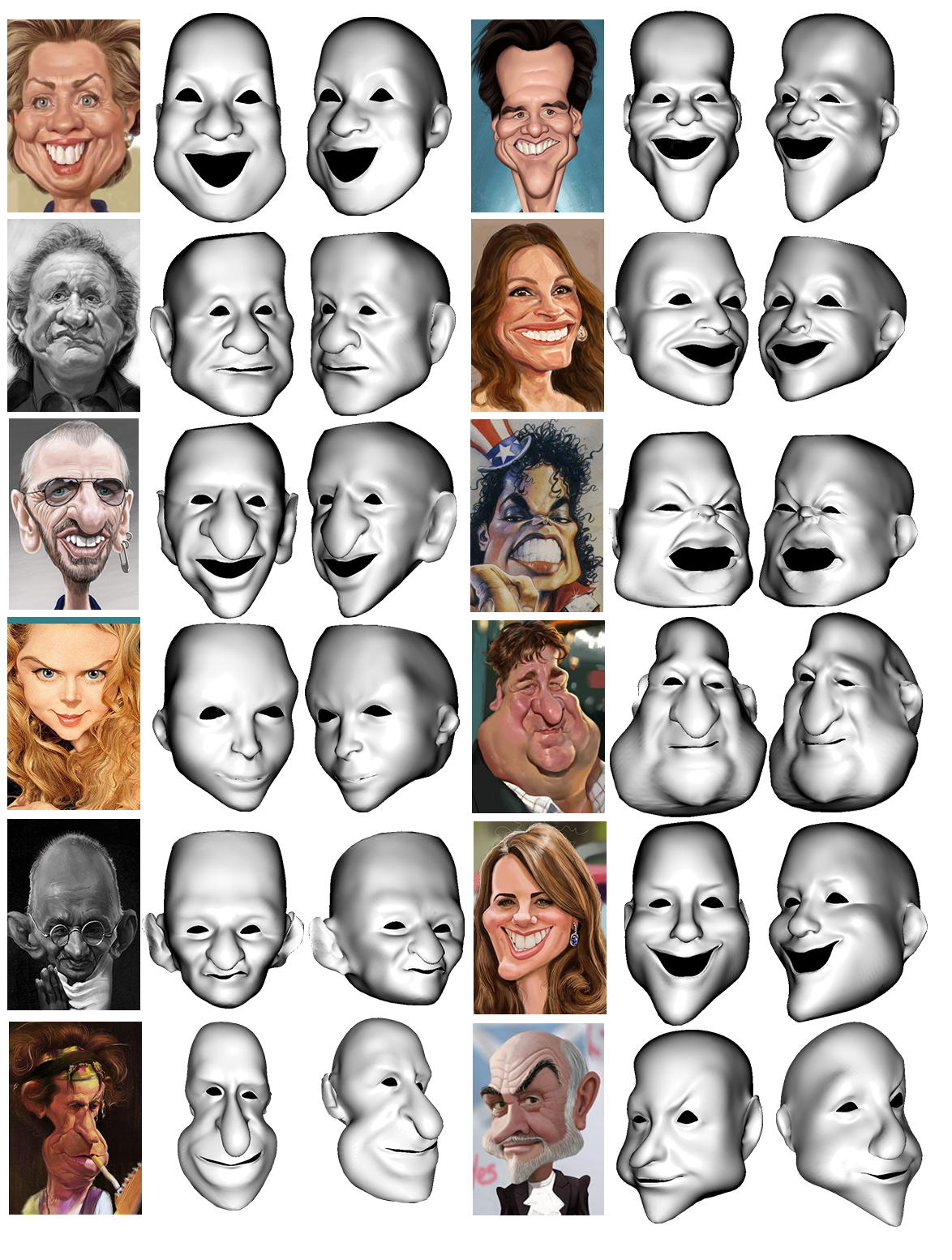}
  \end{center}
  \vspace{-4mm}
  \caption{Results gallery of the proposed framework on \dname{}. The framework has the capability to reconstruct 3D shapes from caricature images with diverse texture and geometry shapes.}
  \label{fig:supp_results_gallery}
  \vspace{-4mm}
\end{figure*}

We present more visual results in Fig.~\ref{fig:supp_results_gallery} to show the effectiveness of our framework. In addition, We show more qualitative results for ablation studies on the framework.
\section{Applications}
\label{sec:suppl_app}

\begin{figure*}
    \begin{center}
    \includegraphics[width=6in]{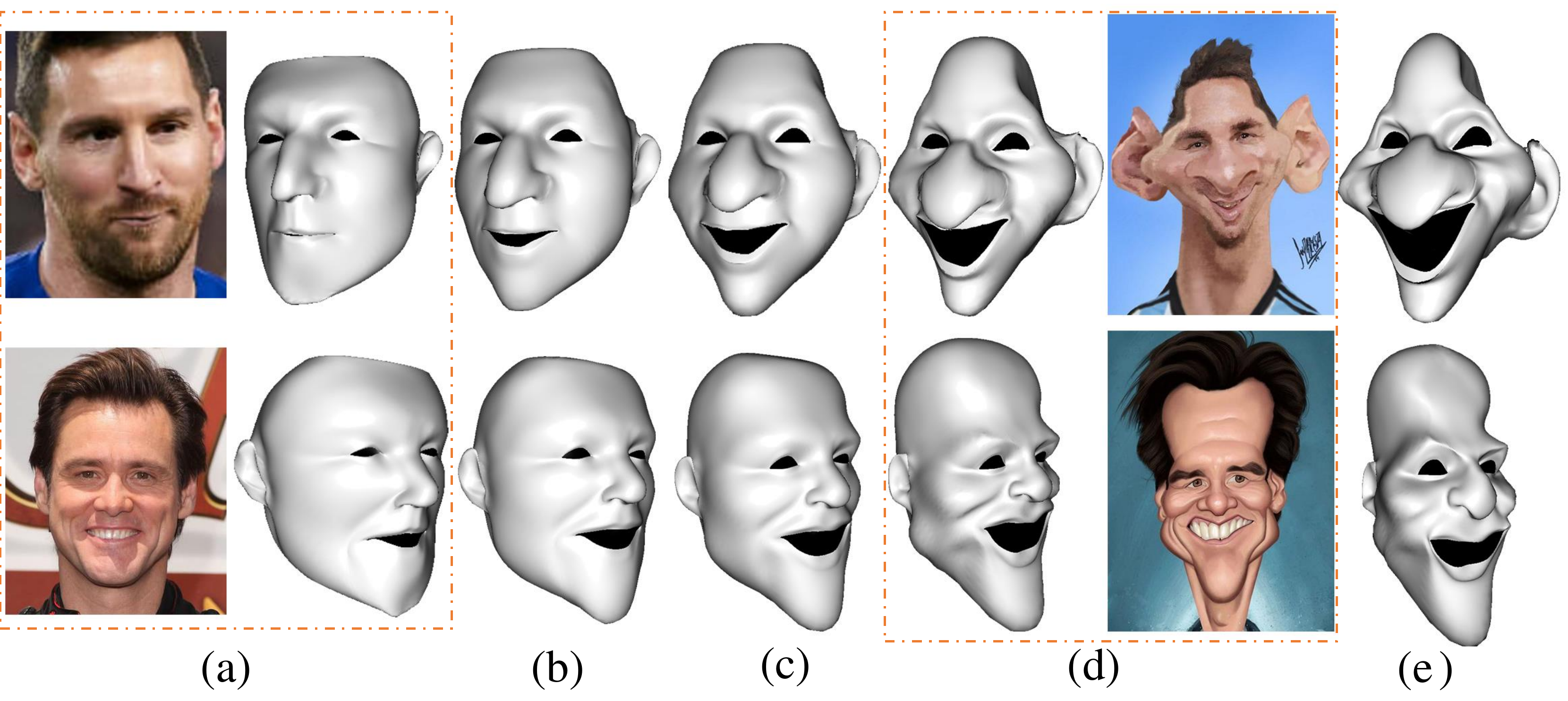}
    \end{center}
  \caption{Samples of interpolation application of our method. Thanks to the uniform topology, it's feasible to generate novel caricature shapes by interpolating the predicted meshes among different inputs. As shown in the figure, (a) are the input real photos and the corresponding reconstructed meshes, (d) are the input caricature images and the generated meshes. (b)(c) are the interpolation results for the meshes from (a)(d). Also, we can perform extrapolating between the meshes (a)(d) to create more exaggerated results, as shown in the last column (e).}
  \label{fig:supp_app}
  \vspace{-2mm}
\end{figure*}

The proposed framework generates caricature meshes with uniform topology. With the well-defined topology, various applications could be developed. In Fig.~\ref{fig:supp_app} we present the mesh generation via interpolating among the predict caricature meshes. 

\begin{figure*}[h]
    \begin{center}
    \includegraphics[width=6in]{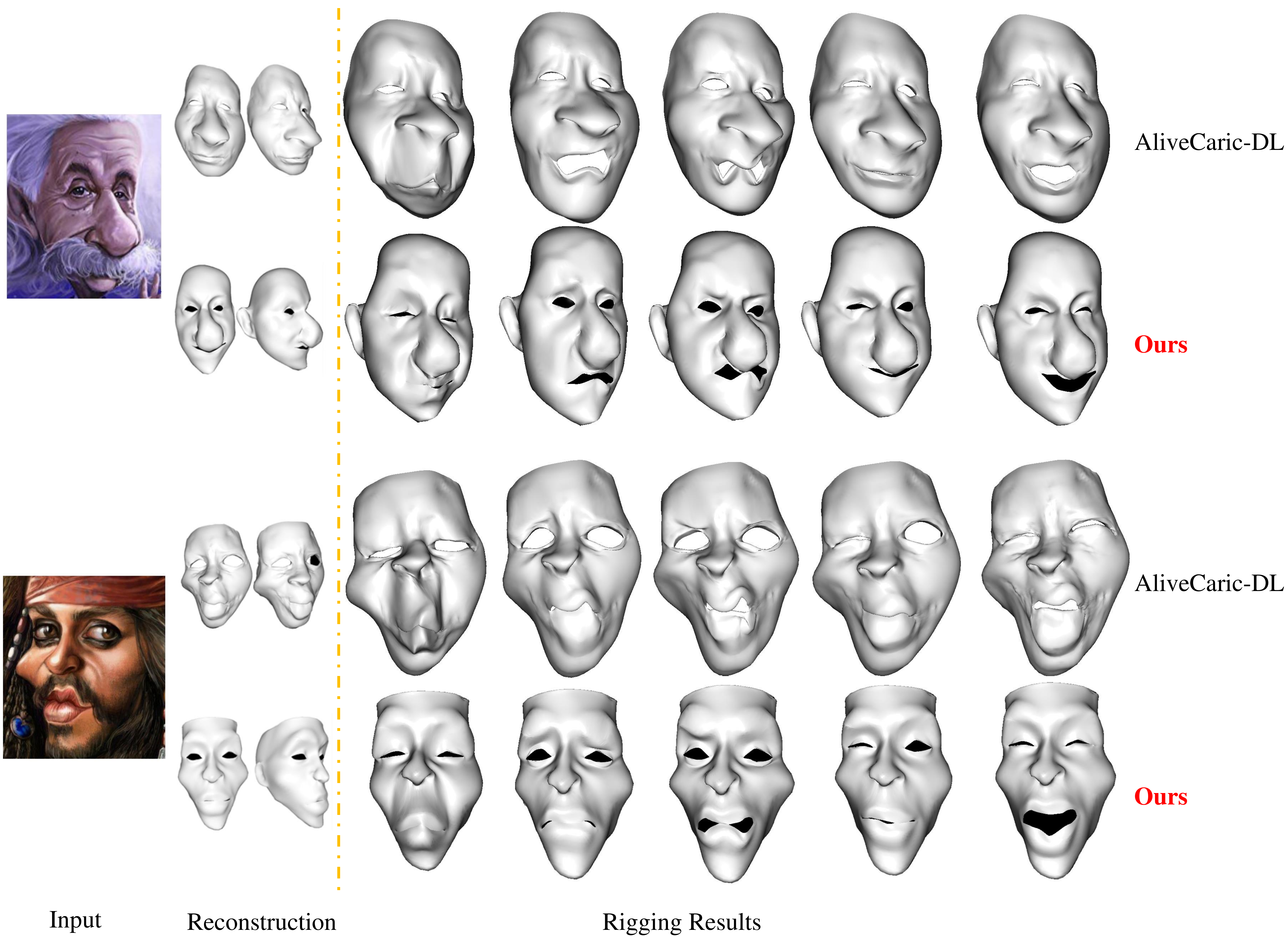}
      \end{center}
  \caption{Rigging samples.}
  \label{fig:rigging}
  \vspace{-6mm}

\end{figure*}

 In Fig.~\ref{fig:rigging}, we compare the rigging results with AliveCaric-DL (ADL). Both results are animated using the same skeleton and skinning weights for fair comparison.
We show that our method supports faithful rigging of our results and preserves better geometric details than ADL.

	\end{document}